\documentclass{article}

\usepackage{arxiv}

\usepackage[utf8]{inputenc} 
\usepackage[T1]{fontenc}    
\usepackage{hyperref}       
\usepackage{url}            
\usepackage{booktabs}       
\usepackage{amsfonts}       
\usepackage{nicefrac}       
\usepackage{microtype}      
\usepackage{lipsum}
\usepackage{wrapfig}
\usepackage[ruled,vlined]{algorithm2e}
\usepackage{float}
\usepackage{microtype}
\usepackage{graphicx}
\usepackage{subfigure}

\usepackage{amsmath,amsfonts,bm}

\def\eqref#1{equation~\ref{#1}}

\def\1{\bm{1}}

\DeclareMathAlphabet{\mathsfit}{\encodingdefault}{\sfdefault}{m}{sl}
\SetMathAlphabet{\mathsfit}{bold}{\encodingdefault}{\sfdefault}{bx}{n}

\DeclareMathOperator*{\argmax}{arg\,max}
\DeclareMathOperator*{\argmin}{arg\,min}

\usepackage[sort&compress,square,comma,authoryear]{natbib}
\DeclareMathOperator{\EX}{\mathbb{E}}

\usepackage{hyperref}

\title{Deep RL With Information Constrained Policies: \\ Generalization in Continuous Control}

\author{
  Tyler Malloy, Chris R. Sims \\
  Department of Cognitive Science\\
  Rensselaer Polytechnic Institute\\
  Troy, NY 12180 \\
  \texttt{\{mallot,simc3\}@rpi.edu} \\
   \And
 Tim Klinger, Miao Liu, Matthew Riemer, Gerald Tesauro \\
  IBM Research AI \\
  Yorktown Heights, NY 10598 \\
  \texttt{\{tklinger, mdriemer, gtesauro\}@us.ibm.com} \\
  \texttt{\{miao.liu1\}@ibm.com}  \\
}

\begin{document}
\maketitle
\begin{abstract}
Biological agents learn and act intelligently in spite of a highly limited capacity to process and store information. Many real-world problems involve continuous control, which represents a difficult task for artificial intelligence agents. In this paper we explore the potential learning advantages a natural constraint on information flow might confer onto artificial agents in continuous control tasks. We focus on the model-free reinforcement learning (RL) setting and formalize our approach in terms of an information-theoretic constraint on the complexity of learned policies. We show that our approach emerges in a principled fashion from the application of rate-distortion theory. We implement a novel Capacity-Limited Actor-Critic (CLAC) algorithm and situate it within a broader family of RL algorithms such as the Soft Actor Critic (SAC) and Mutual Information Reinforcement Learning (MIRL) algorithm. Our experiments using continuous control tasks show that compared to alternative approaches, CLAC offers improvements in generalization between training and modified test environments. This is achieved in the CLAC model while displaying the high sample efficiency of similar methods.
\end{abstract}

\section{Introduction}
The combination of reinforcement learning with deep neural networks (Deep RL) has led to the achievement of significant milestones in AI research \citep[e.g.][]{tesauro1995td, mnih2013playing, silver2016mastering}. However, at the same time, some have called into question the extent to which these approaches are able to demonstrate nontrivial generalization of learning, as compared to simply memorizing state-action sequences \citep{packer2018assessing, cobbe2018quantifying}. In this paper, we look to improve sample efficient generalization in RL by making a connection to the idea of capacity limits in the field of Information Theory. As we will demonstrate, this idea has strong theoretical ties to other recently popular approaches including Maximum Entropy Reinforcement Learning (MERL) \citep{haarnoja2018alg} and KL-regularized RL \citep{tirumala2019exploiting} as well as Mutual-Information Regularized RL (MIRL) \citep{grau2018soft}. Furthermore, our new framework for capacity limited RL motivates a new algorithm called the Capacity Limited Actor Critic (CLAC) that in some environments displays superior out of distribution generalization properties without sacrificing sample efficiency or online learning performance compared to existing methods. 

RL generally consists of a Markov Decision Process (MDP) defined by a state $s$, an action $a$, a transition function $P(s'|s,a)$ that determines the probability of the next state, and a reward function $r(s,a)$. In this setting the primary objective is typically to learn a policy, which is defined as a probability distribution over actions conditioned on states, $\pi(a|s)$. In fact, we would like to learn a policy that optimizes the expected return over horizon $T$ following the state distribution of the policy $\rho_{\pi}$: $J(\pi) = \mathbb{E}_{(s_t,a_t) \sim \rho_{\pi}}[\sum_{t=0}^{T} r(s_t,a_t)]$. 

While MDPs define the setting for an idealized learning agent, any physical communication system (biological or artificial) is necessarily limited to transmitting information at a finite rate. In information theory, the information rate of a channel with input $x \in X$ and output $y \in Y$ is quantified by the so-called \textit{mutual information}, $\mathcal{I}(X,Y)$. For any physical information processing system, $\mathcal{I}(X,Y) \leq \mathcal{C}$ for some finite value of $\mathcal{C}$. Agents with limited information processing capacities should seek to produce behavior that maximizes expected utility. Hence, for any agent, optimal behavior is the result of solving a constrained optimization problem (maximizing the utility of behavior, subject to constraints on available information capacity). Recent work in cognitive science has shown that this constrained optimization perspective on information processing can account for generalization in biological perception \citep{sims2018efficient}. In the present work, we apply this perspective to the reinforcement learning framework. In particular, we consider an agent's policy as an information channel, that maps from the current state of the environment, to a probability distribution over actions, and define optimal behavior subject to a constraint on the information capacity of this channel. More formally, an optimally efficient communication channel is one that minimizes expected loss in utility, $\EX[L(x,y)]$, with respect to this constraint:
\begin{equation*}
\begin{split}
\textbf{Goal: } \text{Minimize} \EX[L(x,y)]  \text{ w.r.t, } p(x|y) \text{, subject to } \mathcal{I}(X,Y) \leq \mathcal{C}
\end{split}
\end{equation*}
This objective is well-studied within the field of Rate Distortion theory \citep{berger1971rate}, a sub-field of information theory. In our application of RD theory onto RL, we consider the policy function $\pi(a|s)$ learned by an RL agent to be a communication channel that maps from the state $s$ onto a probability of performing an action $a$. This allows us to define the optimality condition of an RL agent with a constraint on information representation as
\begin{equation}
\begin{split}
\label{constrained_optimization_mi}
\text{max}_{\pi_{0:T}} \EX_{(s_t,a_t) \sim \rho_{\pi}}\bigg[ \sum_{t=0}^T r(s_t,a_t) \bigg]  \text{ s.t } \EX_{(s_t,a_t) \sim \rho_{\pi}}[\mathcal{I}(\pi(a|s)) \leq \mathcal{C}] \text{  } \forall t
\end{split}
\end{equation}
where $\mathcal{I}(\pi(a|s))$ is the mutual information of the policy function when taken to be the information channel mapping states onto actions. The introduction of this policy mutual information term constrains the agent's policy objective in relation to the information capacity. There are multiple ways of calculating this value which will be discussed in the next section. This allows us to define $\mathcal{C}$, the desired maximum channel capacity, and optimize performance in the environment in relation to this capacity. This optimization can be used to define a learning objective that better reflects the reality of information constraints on physical agents, and as we will see, allows for better control over the trade off of immediate performance and generalization. 

\section{Capacity Limited RL (CLRL)}
\subsection{Capacity-Limited Learning Objective}
Insight from an information-theoretic perspective inspires the goal of maximizing reward obtained subject to some constraints on policy complexity (as measured by its mutual information). The connection between information capacity and generalization in continuous control environments will be motivated more thoroughly in later sections, but the intuitive justification is that policies that are simpler in an information-theoretic sense discourage the use of dissimilar actions in similar states of the environment that only afford a minute improvement in expected reward. In practice, the way we impose a limit on the amount of information that the agent uses to represent its policy is done by applying a penalty to the reward based on this value. This allows us to define a learning objective that regularizes the observed reward:
\begin{equation*} 
J(\pi) = \sum_{t=0}^T \EX_{(s_t,a_t) \sim p_{\pi}}[r(s_t,a_t) - \beta \mathcal{I}(\pi(\cdot|s_t)] 
\end{equation*}
The key difference with the standard RL objective is the added penalty to the reward observed based on the amount of information that would be required to represent the policy. Policies with higher mutual information values have a greater complexity, in an information-theoretic sense, and this weighted value is used to discourage policies that would require a high information capacity channel. Thus, this learning objective will directly encourage the development of policies that are simple (use low information to represent) but have high utility. Additionally, if there are multiple policies that achieve the same performance, this objective will naturally favor the simplest among them. Higher values of $\beta$ skew the learning objective to prefer policies with less required information.

The limitation that is imposed on the information capacity of the agent's policy is introduced by this learning objective. Because this learning objective is used to update the agent's policy throughout training, the information capacity of the learned policy will be dependent on the value of $\beta$. In the extreme, very high values of $\beta$ will train an agent to prefer a policy that requires as little information to represent as possible over any improvement in the reward. Because of the nature of information capacities, this policy could be either uniform in all states of the environment and perform actions randomly everywhere, or deterministic in all states or perform the same action everywhere. Conversely, setting the value of $\beta$ to zero results in the traditional learning objective of maximizing the reward with no limitation on the amount of information utilized by the policy. 

This learning objective is related to Eq \ref{constrained_optimization_mi} due to the fact that for any given information capacity-limit in an environment, there is a corresponding $\beta$ coefficient that induces this limit. In addition, an important feature of this learning objective is that as the $\beta$ parameter changes, it will be optimized by the policy that has the highest performance for that given information capacity. This also means that for an environment with many optimal policies, the learning objective will be optimized by the policy with the lowest information requirement. This is in part the source of the improvement in generalization performance that is afforded by the capacity-limited method, as policies that are over-fit to previous experience typically require more information to represent without achieving a significant increase in performance. This will be demonstrated in the section on the learning environments used to compare performance of the capacity-limited approach with related methods.  
\subsection{Calculating Mutual Information}
The mutual information in the CLRL learning objective can be defined in different ways depending on the features of the learning environment. Because CLRL describes a broad learning objective that can be applied to different existing RL methods, these different methods may be best suited by different approximations of mutual information. For discrete state and action spaces in tabular learning conditions the space is small enough that the mutual information can be defined in terms of the probability mass functions as follows:
\begin{equation}
\mathcal{I}(\pi(a|s)) = \sum_{a \in A} \sum_{s \in S} p_{(s,a)}(s,a) \log \bigg( \dfrac{p_{(s,a)}(s,a)}{p_a(a) p_s(s)} \bigg)
\end{equation}
However, for more complicated problems, especially environments involving continuous state/action spaces, we will need a different approach. One way this can be done is to break up the policy mutual information into the components of its constituent entropies. This calculation of mutual information avoids the approximation of both the marginal state distribution and marginal action distribution, with only one approximation required, being calculated by either of:
\begin{subequations}
\begin{align}
\mathcal{I}(\pi(a|s)) &= \mathcal{H}(\pi_a(a)) - \mathcal{H}(\pi(a|s)) \label{eq:a} \\ 
&= \mathcal{H}(\pi_s(s)) - \mathcal{H}(\pi(s|a)) \label{eq:b} 
\end{align}
\end{subequations}
\subsection{Calculating Marginal Action Distribution}
\label{section:Marginal}
To derive our approximation of the marginal action distribution used in calculating the mutual information, we employ a similar method as used in the MIRL approach, altered to work in continuous action spaces. We begin with the definitions of the stationary distribution over states and actions as described in \cite{grau2018soft}:

\textbf{Definition 1 }(stationary distribution over states) The stationary distribution over states (assumed to exist and to be unique) is defined in vector form as $ \mu_{\pi}^T = \lim_{t \rightarrow \infty} v_0^T P_{\pi}^t $ with $v_0$ being an arbitrary vector of probabilities over states at time $t = 0$. The stationary distribution satisfies $\mu_{\pi}(s') = \sum_{s} P_{\pi}(s'|s) \mu_{\pi}(s)$ and therefore is a fixed point under the state transition probabilities $\mu_{\pi}^T = \mu_{\pi}^T P_{\pi}$.

\textbf{Definition 2 }(stationary distribution over actions)\textit{ Let $\mu_{\pi}(s)$ be the stationary distribution over states induced by the policy $\pi$. Then the stationary distribution over actions under $\pi$ is defined as $\rho_{\pi}(a) := \sum_{s \in S} \mu_{\pi}(s) \pi(a|s)$}.

To determine the marginal action distribution $p_{\pi}(a)$ we begin with the definition of the stationary distribution over actions. We are interested in applications with continuous action spaces and tractable polices, and thus limit the continuous policy function to a set of functions $\Pi$ defined by the spherical Gaussian, as is done in the Soft-Actor Critic \cite{haarnoja2018soft}. This allows us to define the marginal action distribution similarly as $\mathcal{N}(\mu_{\rho}, \sigma_{\rho}^2)$ and calculate the mean and variance as:
\begin{equation*}
\begin{split}
\mu_{\rho} & = \sum_s p(s) \mu_s  \\
\sigma_{\rho}^2 & = \sum_s p(s) \sigma_s^2 + \sum_s p(s) \big( \mu_s \big)^2 - \bigg( \sum_s p(s) \mu_s \bigg)^2
 \end{split}
\end{equation*}
Where $\mu_s$ a vector representing the mean of the policy at state s with the same dimension as the action space, and $\sigma_s^2$ is the variance. As our policy is stochastic, at each step $n$ through the environment our policy outputs a mean $\mu_n$ and variance $\sigma^2_n$ defining the action that will be performed in that state. This gives us the online updating method for the estimate of the marginal action distribution variance $\hat{\sigma}_n^2$ and mean $\hat{\mu}_n$ with the learning rate $\alpha$.
\begin{equation*}
\begin{split}
\hat{\mu}_n      & = \alpha \mu_n + (1-\alpha) \hat{\mu}_{n-1} \\
\hat{\sigma}_n^2 & = \alpha \sigma_n^2 + (1-\alpha) \hat{\sigma}_{n-1}^2 
                 + \big( \alpha \mu_n^2 + (1-\alpha) (\hat{\mu}_{n-1})^2 \big)  
                 - \big( \alpha \mu_n + (1-\alpha) \hat{\mu}_{n-1}  \big)^2  
 \end{split}
\end{equation*}
Due to the definition of the stationary distributions over states $\mu_{\pi}(s)$ and actions $\mu_{\pi}(a)$, we can show that this approximation approaches the true mean and variance of $p_{\pi}(a)$ as the distribution of the states in our experience approaches the steady state distribution (a full derivation is provided in the supplementary material). The choice of estimating the current marginal action distribution with a Gaussian admixture as opposed to alternative approaches such as an additional neural network is done due to the mathematical justification that in the limit of experience this approximation is guaranteed to approach the true marginal. 

\section{Related Work}
The Capacity Limited algorithm is situated in a family of regularized Markov Decision Process (MDP) algorithms. Within this group of regularized MDP algorithms is a class defined by a regularization based on the entropy of the agent's policy \cite{neu2017unified}. In this section we contrast our approach to similar methods within this family, differentiating them based on their motivations and methods. 

\subsection{Connections to Maximum Entropy Reinforcement Learning (MERL)}
Maximum Entropy RL \citep{haarnoja2018soft} is a popular RL framework that also makes use of a regularized learning objective to maximize:
\begin{equation*}
\begin{split}
J(\pi) &= \sum_{t=0}^T \EX_{(s_t,a_t) \sim \rho_{\pi}}[r(s_t,a_t) + \alpha \mathcal{H}(\pi(\cdot|s_t))]
\end{split}
\end{equation*}
This learning objective alters the traditional method by augmenting the reward maximization objective with an additional weighted value based on the entropy of the policy $\alpha \mathcal{H}(\pi(\cdot|s_t))$. This encourages policies that are closer to the uniform distribution and thus more random, since these action distributions will have a higher entropy. The entropy coefficient $\alpha$ controls the weight balancing the reward and entropy of the policy. As this value increases to infinity, the model will learn to always prefer policies that are completely uniform. When this value is 0, the model will only consider the reward returned by the environment and is equivalent to the traditional RL learning objective. 

As there is a close theoretical tie between the frameworks, we leverage the successful Maximum Entropy method Soft-Actor-Critic \citep[SAC, ][]{haarnoja2018soft} as the main point of comparison for our approach. SAC is an off-policy RL method that displays high sample efficiency in continuous control environments \citep{haarnoja2018alg}. To make the comparison easier with SAC, we utilize the \eqref{eq:a} for the calculation of $\mathcal{I}(\pi(a|s))$. One benefit to this approach is that any MERL method will naturally need to compute the value $\mathcal{H}(\pi(a|s))$ and can be altered with relative ease to include the additional marginal action term, as is done in this implementation. In the next section on the Deep CLAC algorithm, we apply this alteration to the existing Soft-Actor Critic algorithm that uses the MERL objective to improve exploration. This is done to show that the Capacity-Limited objective can be used to alter existing methods with relative ease, especially ones that already compute the policy entropy $\mathcal{H}(\pi(a|s))$, such as MERL methods. 

The CLRL and MERL approaches are similar in that they use learning objectives that maximize a regularized reward. Although the CLRL learning objective penalizes policies with higher mutual information $\mathcal{I}(\pi(a|s))$ as opposed to rewarding ones with lower entropy. These two values are related by the equation we use to calculate the mutual information $\mathcal{I}(\pi(a|s)) = \mathcal{H}(\pi_a(a)) - \mathcal{H}(\pi(a|s))$. The intuition behind the CLRL approach is not merely altering this objective by introducing a penalty to $\mathcal{H}(\pi_a(a))$ but rather in preferring a learning objective that produces policies that require less information to represent. As we will see in results from experimentation, this learning objective often results in policies that are better suited towards the goal of generalization as we will define it. 

\subsection{Connections to KL-Regularized Reinforcement Learning}
Recent work \citep{tirumala2019exploiting} has furthered the justification of using the KL-divergence as a regularizer on learned policies. In this section we explore the connection to CLRL.

The KL-Regularized expected reward objective maximizes
\begin{equation*}
\begin{split}
J(\pi) & = \sum_{t=0}^T \EX_{(s_t,a_t) \sim p_{\pi}}[r(s_t,a_t) - \alpha\text{KL}(a_t|x_t)] 
\\
& = \sum_{t=0}^T \EX_{(s_t,a_t) \sim p_{\pi}}\Bigg[r(s_t,a_t) - \alpha \log\dfrac{\pi(a_t|s_t)}{\pi_0(a_t|s_t)} \Bigg] \\
\end{split}
\end{equation*}
referring to the full history up to time t as $x_t = [s_0,a_0,...s_t,a_t]$. This method is used in the hierarchical learning domain which separates broad high-level task focused decisions and specific low-level task agnostic actions \citep{tirumala2019exploiting}. In KL-RL, this is done by training a task specific policy $\pi(a|s)$ as well as a default policy $\pi_0(a|s)$ and using the KL divergence between these two as a regularizer. In this way the agent learns to generalize across tasks by increasing the reward observed when reusing learned behaviour across tasks. Because CLAC is not hierarchical in nature we do not compare the performance of KL-RL to CLAC, but we do relate them theoretically. 

As noted in \cite{tirumala2019exploiting} the maximum entropy objective is a special case of the KL-regularized objective where the default policy is taken to be a uniform distribution. This is intuitively in line with the motivation behind the MERL objective, as it seeks to increase exploration by encouraging random behavior where possible. A similar relation can be made between KL-RL and the CLRL defined here, where the `default policy' is taken to be the marginal action distribution. The coefficient $\alpha$ is renamed to $\beta$ to correspond with the rate-distortion theory notation. This derivation relating KL-RL to CLRL is given by first taking the definition of the KL-RL objective:
\begin{align*}
J(\pi) &= \sum_{t=0}^T \EX_{(s_t,a_t) \sim p_{\pi}}[r(s_t,a_t) - \alpha\text{KL}(a_t|x_t)] \\ 
J(\pi) &= \sum_{t=0}^T \EX_{(s_t,a_t) \sim p_{\pi}}\bigg[r(s_t,a_t) - \beta \log\dfrac{\pi(a_t|s_t)}{\pi_0(a_t|s_t)} \bigg] \\
\end{align*}
We can write the denominator of this logarithm term generally as a function of a state and action such as $\rho(a,s)$. This allows us to define KL-RL to be the case where  $\rho_0(a,s) = \pi_0(a_t|s_t)$. In this same manner SAC can be defined as the uniform distribution $\rho(a,s) = 1/||A||$.  Finally, CLAC can be defined as $\rho(a,s) = \pi_a(a)$ the marginal action probability  (a full derivation is provided in the supplementary material).
\begin{align*} J(\pi) &= \sum_{t=0}^T \mathbb{E}_{(s_t,a_t) \sim p_{\pi}}\big[r(s_t,a_t)\big] - \beta \mathcal{I}(\pi(a_t|s_t)) \\ \mathcal{I}&(\pi(a_t|s_t))= \mathbb{E}_{s_t,a_t}\bigg[ - \log\bigg[\dfrac{\pi(a_t|s_t)}{\pi_a(a_t)} \bigg]\bigg] \\ \end{align*}
As with the relation between SAC and KL-RL, this relation is in line with the motivation behind CLAC, specifically in encouraging the reuse of policies across similar states, learning a useful prior, and learning a policy that requires less information to represent. 

\subsection{Connections to Mutual Information Reinforcement Learning}
The CLRL approach is similar to the Mutual Information Reinforcement Learning (MIRL) model \cite{grau2018soft}, which uses the mutual information between states and actions as a regularizer in soft Q-learning. The main difference between that approach and ours is that in MIRL the focus is on improving exploration rather than learning simplified policies, so there is no limitation on the amount of information that can be used by the agent to represent the policy. MIRL uses a traditional $\epsilon$-greedy strategy of determining which action to take given some random uniformly distributed $u$ as: 
\begin{equation*}
a_i = \begin{cases} \argmax_a \pi_i(a|s_i) & \mbox{if } u > \epsilon \\ a \sim \rho(a) & \mbox{if } u \leq \epsilon \end{cases} 
\end{equation*}
This is different from the Capacity-Limited approach which uses the same policy at each step in the environment. Additionally, the parameter controlling the contribution of the mutual information to the learning objective is decreased over time in MIRL. The reason for this is that the mutual information term is meant to alter the exploration method of the agent, and not the final policy that it learns. This motivation is significantly different from that of the Capacity-Limited approach which seeks to improve the generalizability of the learned policy by penalizing policies that require a large amount of information to represent without a proportionally large improvement in the expected reward. 

\section{Deep Capacity Limited Actor Critic Learning}
There are many RL approaches that could be applied in optimizing the capacity-limited learning objective. Here, we implement the CLAC algorithm based on a baseline implementation \citep{stable-baselines} of the Soft Actor-Critic as described in \citep{haarnoja2018soft}. Although the baseline implementation package does not include an MIRL implementation, the main difference can be effectively created by taking our existing CLAC method and iteratively decreasing the mutual information coefficient $\beta$ and adding an $\epsilon$-greedy value for taking random actions. This altered version of CLAC will serve as a comparison of the MIRL method against SAC and CLAC in the following results sections. Although this version of MIRL may differ slightly from the original implementation there is an added benefit in terms of comparison in that all other model features besides the relevant differences will be the same between the MIRL and CLAC models.  

The reason for basing our CLAC on the existing SAC implementation is to allow us to compare the CLRL and a MERL objectives while keeping other model features static. Specifically, this network structure uses policy iteration consisting of a value network with target network updating, 2 q-function networks, and a policy network. The derivation of SAC relies on the so-called `soft value function' $V(s_t) = E_{a_t \sim \pi}[Q(s_t,a_t) + \alpha \mathcal{H}(\pi(a_t|s_t)]$. Here we use the analogous capacity-limited value function $V(s_t) = E_{a_t \sim \pi}[Q(s_t,a_t) - \beta \mathcal{I}(\pi(a_t|s_t)]$. Because we choose to define the mutual information value in terms of the constituent entropies, a slight alteration to SAC method is carried forward into defining the gradients of the policy, q-network and value network. This allows us to define the gradients used to update these three networks as follows (a full derivation as well as a CLAC algorithm are provided in the supplementary material)
\begin{equation*}
\begin{split}
\hat{\nabla}_{\chi} J_V(\chi) & = \nabla_{\chi} V_{\chi}(s_t)(V_{\chi}(s_t) - Q_{\kappa}(s_t,a_t)  - \beta (\log(\pi_{\mu}(a_t)) - \log(\pi_{\phi}(s_t,a_t)))) \\
\hat{\nabla}_{\kappa} J_Q(\kappa) & = \nabla_{\kappa} Q_{\kappa}(s_t,a_t)(Q_{\kappa}(s_t,a_t)  - r(s_t,a_t) - \gamma V_{\hat{\chi}}(s_{t+1})) \\
\hat{\nabla}_{\phi} J_{\pi}(\phi) & = \nabla_{\phi} \log \pi_{\phi}(a_t|s_t) + \big( \nabla_{a_t} \beta (\log \pi_{\mu}(a_t)  - \log \pi_{\phi}(a_t|s_t)) - \nabla_{a_t} Q(a_t,s_t) \big) \nabla_{\phi}f_{\phi}(\eta_t;s_t)
\end{split}
\end{equation*}

where $f_{\phi}(\eta_t;s_t)$ defines the policy under the same reparameterization trick $a_t = f_{\phi}(\eta_t;s_t)$, used in SAC, where $\eta_t$ is a noise vector sampled from a spherical Gaussian distribution \citep{haarnoja2018soft}. The current approximation of the mean $\hat{\mu}_n$ and standard deviation $\hat{\sigma}_n^2$ of the marginal action distribution $\mathcal{N}(\hat{\mu}_n, \hat{\sigma}_n^2)$ are updated at each environment step based on the mean and variance of the action performed by the agent, according to the equation described in section \ref{section:Marginal}. 

Using a different approach to estimating the marginal action probability would be possible, such as training an additional DNN to estimate this marginal and updating it based on observed actions. This would replace the batch approximation step with a gradient update of the marginal action distribution approximation network: $\mu \leftarrow \mu - \lambda_M \hat{\nabla}_{\mu} J_M(\mu)$. 

This algorithm utilizes a fixed mutual information coefficient $\beta$ that is used to balance the contribution of the policy mutual information and the reward in the learning objective. Another option that is discussed in depth in section \ref{section:Auto} is to automatically update the mutual information coefficient based on some desired information capacity for the policy channel. For this method to be used, the above algorithm is simply augmented to include the coefficient update step $\beta \leftarrow \beta - \gamma \hat{\nabla}_{\beta}J(\beta)$ at each gradient step. Together these give the capacity-limited actor-critic algorithm:

\begin{algorithm}[H]
    Initialize: parameter vectors $\phi$, $\chi$, $\kappa_1$, $\kappa_2$ \\
    Initialize: Memory $\mathcal{D} = \emptyset$ \\
    Initialize: Learning Rates: $\lambda_V$, $\lambda_Q$, $\lambda_{\pi}$, $\alpha$, $\tau$ \\
    Initialize: marginal action mean $\hat{\mu}_0 = \textbf{0}$ \\
    Initialize: marginal action variance:  $\hat{\sigma}^2_0 = \textbf{0}_{n,n}$ \\
    \For{each iteration}{
        \For{each environment step}{
        $a_t \sim \pi_{\phi}(a_t|s_t)$ \\
        $s_{t+1} \sim p(s_{t+1} | s_t, a_t)$ \\
        $\hat{\mu}_n \leftarrow \alpha \mu_n + (1-\alpha) \hat{\mu}_{n-1}$ \\
        $\hat{\sigma}_n^2 \leftarrow \alpha \sigma_n^2 + (1-\alpha) \hat{\sigma}_{n-1}^2
                 + \big( \alpha \mu_n^2 + (1-\alpha) (\hat{\mu}_{n-1})^2 \big)$  \\
                 $- \big( \alpha \mu_n + (1-\alpha) \hat{\mu}_{n-1}  \big)^2$   \\
        $\pi_{\mu}(a_t) \sim \mathcal{N}(\hat{\mu}_n, \hat{\sigma}_n^2)$ \\ 
        }
        \For{each gradient step}{
        $\chi \leftarrow \chi - \lambda_V\hat{\nabla_{\chi}} J_V(\chi) $ \\
        $\kappa_i \leftarrow \kappa_i - \lambda_Q \hat{\nabla_{\kappa_i}} J_Q(\kappa_i)  $ for $i \in {1,2}$ \\
        $\phi \leftarrow \phi - \lambda_{\pi} \hat{\nabla_{\phi}} J_{\pi}(\phi)$ \\
        $\bar{\chi} \leftarrow \tau \chi + (1-\tau) \bar{\chi}$ \\
        }
    }
\caption{Capacity-Limited Actor-Critic\label{IR}}
\end{algorithm}

\section{Learning Environments}
\subsection{Continuous N-Chain}
\label{section:N-chain-generalization}

We explore a continuous action space version of the n-chain environment as described by \cite{strens2000bayesian}. This environment consists of $N$ states with agents starting in the $S_1$ state and $S_N$ as the terminal state. Agents act by selecting a continuous value from [0,1], and the probability of them moving to the next state, $p(s_{t+1}|a_t,s_t)$, is proportional to the difference between their action and the hidden value H depending on the state they are in, $H_{s_t}$. These hidden values are sampled from a Beta distribution $Beta(a,b)$. The reward is -1 for all states and 0 for the final state. For all tests shown here the number of states $N = 5$, and beta distribution shape parameters $a = 10$, $b = 25$. In the supplementary material we show and graph the state transition function and a diagram for the state and hidden values sampled from the beta distribution.

\begin{figure}[htbp!]
\begin{centering}
    \includegraphics[width=1.0\textwidth]{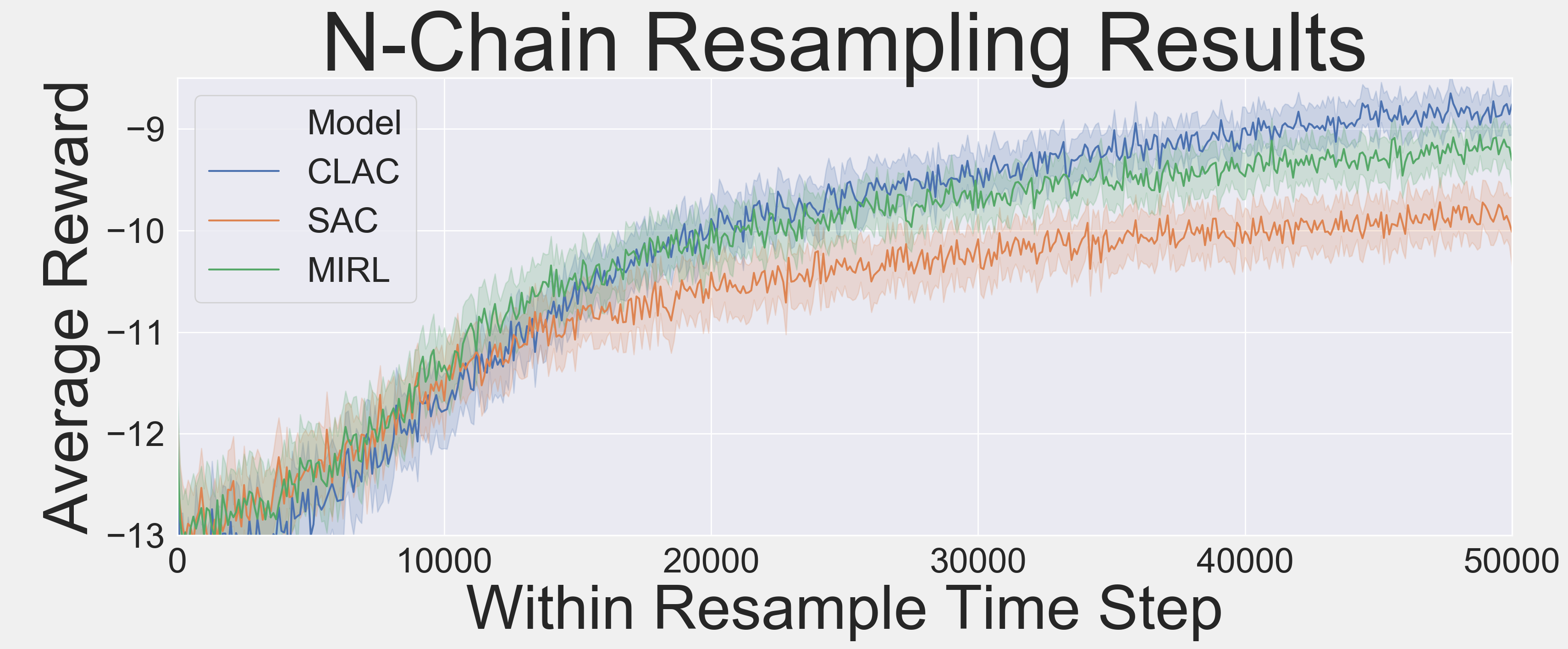}
    \caption{Average reward by time step from 48 agents trained on 5 re-samplings of 50K steps through the N-chain environment with the number of states as n=5 and a Beta(10,25) distribution sampling hidden values. Shaded regions represent standard deviation. Hidden values are re-sampled after each set of 50K steps. All models used the same initial coefficient 0.5}
    \label{fig:ResamplePerformance}
\end{centering}
\end{figure}

One measure of the robustness and generalizability of an agent's policy in the continuous n-chain environment is how quickly it can learn when the hidden values that it is trained on are altered. In this set of tests we train a SAC, CLAC, and MIRL agents on the same hidden values for 50K steps before re-sampling and repeating this process for a total of 1M steps through the environment. Figure \ref{fig:ResamplePerformance} shows the performance results within the final re-sample of hidden values after the series of training on the same series of re-sampled hidden values for pairs of each 48 agents. 

In Figure \ref{fig:ResamplePerformance} we can see that the performance of the CLAC model outperforms both the SAC and MIRL based version of CLAC which decreases the value of the beta coefficient throughout training. Although the CLAC model has lower performance immediately after the hidden values are resampled, it quickly improves its performance and approaches the optimal performance in this task. This indicates that the CLAC model has learned a useful prior over the optimal actions to perform by shifting the actions taken towards the underlying beta distribution used to generate the hidden values. 

These results show that imposing a limitation to the information capacity of an agent's policy results in behaviour that is more robust to changes in this learning environment. This environment is specifically designed to showcase the difference between the behaviour of these models, and we believe that it represents some of the challenges in generalization present in many naturalistic tasks such as continuous control in real world and simulated robotics.

\subsection{Robot Control Environments}
We are interested in determining the potential benefit of a capacity-limited approach in more complex learning environments. In order to determine this, we compared performance of CLAC and SAC agents in the more complex Inverted Double Pendulum and Ant Walker learning environment within the pybullet physics simulation environment \citep{coumans2019}. To compare the generalizability of these two models we employ a similar method as described in \cite{packer2018assessing} with a uniform re-sampling of environment parameters in the randomization case and a sampling from a disjoint set in the extreme case. We randomize the robot joint friction, body mass, applied force, and simulated gravity. A complete list of parameters and ranges is included in the supplementary material.

\begin{figure}[htbp!]
    \begin{centering}
    \includegraphics[width=1.0\textwidth]{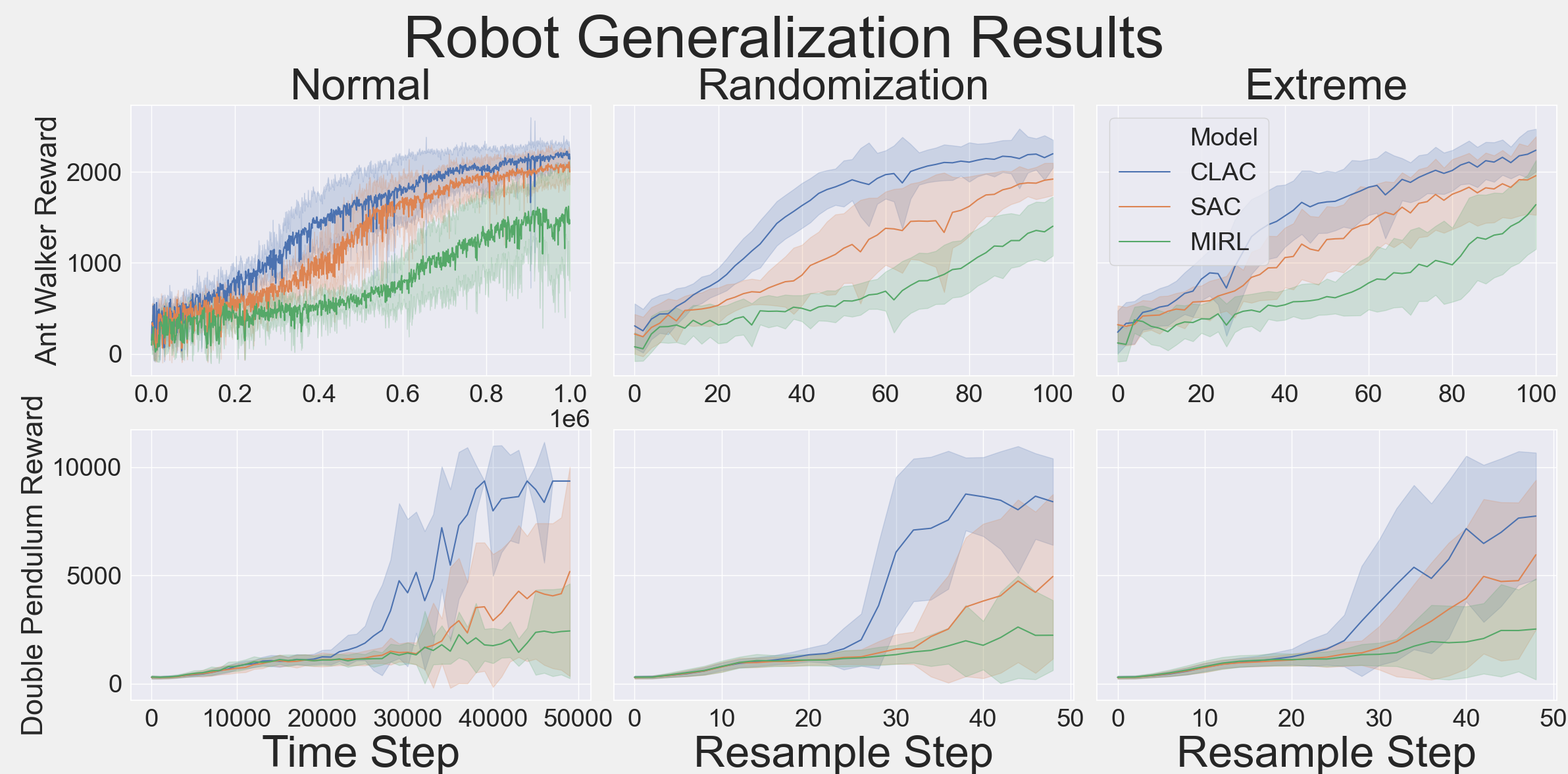}
    \caption{Top: Ant walker controller task. Bottom: Double Pendulum balancing task. Left: Non-randomized training results of 1M (Ant) and 50K (Pendulum) time steps with 8 walker and 20 pendulum agents. Middle: Generalization results with environment parameters re-sampled from a uniform [95-105\%] 100 or 50 times. Right: Generalization results with environment parameters re-sampled from a disjoint set [90-95\%] and [105-110\%] 100 or 50 times. Error bars represent standard deviation. Best performing coefficients for the non-randomized task (left) sampled from [0-2.0] in 0.05 windows and reused for all tests.}
    \end{centering}
\label{fig:AntResults}
\end{figure}

These results indicate that the improvement in generalization performance afforded by the CLAC method is retained in more complex environments with higher dimension action spaces. In both the randomized and extreme randomized conditions, the CLAC agent was able to retain it's high performance while the SAC and MIRL models were significantly impacted by changing environment parameters. Comparing results from the Inverted Double Pendulum task and the Ant Walker task, we can see a clearer improvement in generalization performance in the latter task. This is promising as the ant walker task is more demanding, and may indicate that future environments with more complex tasks may be better suited for generalization through a capacity-limited approach, although future research would be necessary to fully understand which environments can be expected to improve generalization through this method. 

We believe that this type of generalization and robustness to changing physical parameters represents an important aspect of real-world robotics. The types of parameters changes here such as the mass of a body piece or the friction of a joint are emblematic of the types of changes a real agent may have to account for. When a human picks up a coffee cup, they experience a change in the weight they are carrying on that limb, as well as an altered friction in their joints. For RL agents to be deployed in real-world environments, a strong robustness to these types of changes is important. 

\section{Automatic Mutual Information Coefficient Adjustment} \label{section:Auto}
In defining the algorithm for encouraging policies that required less information to represent, we took the practical approach of applying a penalty to the reward observed by an agent based on the complexity of the agents policy. However, there is a way that we can regain our original objective of maximizing reward subject to a given policy channel capacity. This is done by defining a target mutual information for our policy, and automatically updating our coefficient to approach that desired target while training. This is done much in the same way as the automatic adjustment of the entropy coefficient $\alpha$ used in SAC that is described in \cite{haarnoja2018alg}. A full derivation of this method is provided in the Appendix. This derivation provides us the following gradient used to update the value of $\beta$:

\begin{equation*}
J(\beta) = \EX_{a_t \sim \pi_t}[\beta (\log \pi_t(a_t|s_t) - \log \pi_t(a_t|s_t)) + \beta \bar{\mathcal{C}}]
\end{equation*}
where $\mathcal{C}$ is the desired maximum policy channel capacity. Thus we can define the automatic mutual information coefficient adjustment version of Algorithm 1, as including the update:
\begin{equation*}
\beta \leftarrow \beta - \gamma \hat{\nabla}_{\beta}J(\beta)
\end{equation*}

As we have seen, the mutual information coefficient controls the extent to which the CLRL algorithm prioritizes some degree of policy generalizability over training reward maximization. However, as with the entropy coefficient in SAC, this mutual information coefficient is sensitive to the scale of the reward of the learning environment, because the penalty $-\beta \mathcal{I}(\pi(a|s))$ is applied to the reward observed in the environment. Automatically adjusting the entropy coefficient $\alpha$ is partly motivated by invariance to reward scale, and the same benefit is gained by automatically adjusting the mutual information coefficient in CLAC.

However with CLAC, this automatic adjustment has the additional connection to the original motivation for applying a capacity-limit, to allow us to define a channel capacity and maximize reward relative to it. Another benefit of this approach is mitigating some of the potential issues caused by negative transfer, a difficult open question in the area of generalization where attempting to transfer learning can negatively impact performance (\cite{taylor2009transfer}). These negative effects could be mitigated by having the model update the $\beta$ coefficient throughout learning and determine which value of the coefficient best reflects the degree to which knowledge of a policy in one state can be applied to other states. Only tight constraints on information capacity should be disastrously impacted by negative transfer, as weaker constraints do not force agents to reuse policies in states where they are not associated with rewards. 

\section{Discussion}
In this work we present a formalization of the Capacity-Limited Reinforcement Learning (CLRL) objective and relate it to existing methods in Deep Reinforcement Learning. We argue that CLRL defines a broad approach that can be applied onto existing RL methods to improve some aspects of generalization that are particularly relevant for issues like continuous control. To support this position we present the the Capacity-Limited Actor-Critic (CLAC), an application of CLRL onto a deep off-policy actor-critic model. This model is developed based on a high performing method in continuous control for robotics applications, and thus is compared to this previous model as well as another model that shares a similar learning objective. We use a continuous N-chain and continuous control environments to clarify the impact on performance and generalization afforded by the CLAC model.

Differences in performance when altering environment parameters in continuous control environments show improved generalization through optimizing the CLRL learning objective over the MERL method. This is a key result as it clarifies the difference between the learning objectives that are being optimized in these methods. MERL based methods use an entropy coefficient to encourage exploration through more random behaviour where possible. MIRL uses the policy mutual information term to regularize the MERL objective, deteriorating the relevance over time. Meanwhile, CLRL based methods use the mutual information coefficient to balance one aspect of generalization, and is motivated by the natural capacity for storing and processing information that exists in biological agents. This difference present in CLAC enabled it to outperform existing methods in some environments with perturbed features, displaying the improvement in generalization afforded by a capacity-limited learning objective.

\newpage 

\bibliography{arxiv}

\begin{thebibliography}{16}
\providecommand{\natexlab}[1]{#1}
\providecommand{\url}[1]{\texttt{#1}}
\expandafter\ifx\csname urlstyle\endcsname\relax
  \providecommand{\doi}[1]{doi: #1}\else
  \providecommand{\doi}{doi: \begingroup \urlstyle{rm}\Url}\fi

\bibitem[Berger(1971)]{berger1971rate}
Berger, T.
\newblock \emph{Rate Distortion Theory: A Mathematical Basis for Data
  Compression}.
\newblock Prentice Hall, 1971.

\bibitem[Cobbe et~al.(2018)Cobbe, Klimov, Hesse, Kim, and
  Schulman]{cobbe2018quantifying}
Cobbe, K., Klimov, O., Hesse, C., Kim, T., and Schulman, J.
\newblock Quantifying generalization in reinforcement learning.
\newblock \emph{arXiv preprint arXiv:1812.02341}, 2018.

\bibitem[Coumans \& Bai(2016)Coumans and Bai]{coumans2019}
Coumans, E. and Bai, Y.
\newblock Pybullet, a python module for physics simulation for games, robotics
  and machine learning.
\newblock \url{http://pybullet.org}, 2016.

\bibitem[Grau-Moya et~al.(2018)Grau-Moya, Leibfried, and Vrancx]{grau2018soft}
Grau-Moya, J., Leibfried, F., and Vrancx, P.
\newblock Soft q-learning with mutual-information regularization.
\newblock In \emph{International Conference on Learning Representations}, 2018.

\bibitem[Haarnoja et~al.(2018{\natexlab{a}})Haarnoja, Zhou, Abbeel, and
  Levine]{haarnoja2018soft}
Haarnoja, T., Zhou, A., Abbeel, P., and Levine, S.
\newblock Soft actor-critic: Off-policy maximum entropy deep reinforcement
  learning with a stochastic actor.
\newblock \emph{arXiv preprint arXiv:1801.01290}, 2018{\natexlab{a}}.

\bibitem[Haarnoja et~al.(2018{\natexlab{b}})Haarnoja, Zhou, Hartikainen,
  Tucker, Ha, Tan, Kumar, Zhu, Gupta, Abbeel, et~al.]{haarnoja2018alg}
Haarnoja, T., Zhou, A., Hartikainen, K., Tucker, G., Ha, S., Tan, J., Kumar,
  V., Zhu, H., Gupta, A., Abbeel, P., et~al.
\newblock Soft actor-critic algorithms and applications.
\newblock \emph{arXiv preprint arXiv:1812.05905}, 2018{\natexlab{b}}.

\bibitem[Hill et~al.(2018)Hill, Raffin, Ernestus, Gleave, Traore, Dhariwal,
  Hesse, Klimov, Nichol, Plappert, Radford, Schulman, Sidor, and
  Wu]{stable-baselines}
Hill, A., Raffin, A., Ernestus, M., Gleave, A., Traore, R., Dhariwal, P.,
  Hesse, C., Klimov, O., Nichol, A., Plappert, M., Radford, A., Schulman, J.,
  Sidor, S., and Wu, Y.
\newblock Stable baselines.
\newblock \url{https://github.com/hill-a/stable-baselines}, 2018.

\bibitem[Mnih et~al.(2013)Mnih, Kavukcuoglu, Silver, Graves, Antonoglou,
  Wierstra, and Riedmiller]{mnih2013playing}
Mnih, V., Kavukcuoglu, K., Silver, D., Graves, A., Antonoglou, I., Wierstra,
  D., and Riedmiller, M.
\newblock Playing atari with deep reinforcement learning.
\newblock \emph{arXiv preprint arXiv:1312.5602}, 2013.

\bibitem[Neu et~al.(2017)Neu, Jonsson, and G{\'o}mez]{neu2017unified}
Neu, G., Jonsson, A., and G{\'o}mez, V.
\newblock A unified view of entropy-regularized markov decision processes.
\newblock \emph{arXiv preprint arXiv:1705.07798}, 2017.

\bibitem[Packer et~al.(2018)Packer, Gao, Kos, Kr{\"a}henb{\"u}hl, Koltun, and
  Song]{packer2018assessing}
Packer, C., Gao, K., Kos, J., Kr{\"a}henb{\"u}hl, P., Koltun, V., and Song, D.
\newblock Assessing generalization in deep reinforcement learning.
\newblock \emph{arXiv preprint arXiv:1810.12282}, 2018.

\bibitem[Silver et~al.(2016)Silver, Huang, Maddison, Guez, Sifre, Van
  Den~Driessche, Schrittwieser, Antonoglou, Panneershelvam, Lanctot,
  et~al.]{silver2016mastering}
Silver, D., Huang, A., Maddison, C.~J., Guez, A., Sifre, L., Van Den~Driessche,
  G., Schrittwieser, J., Antonoglou, I., Panneershelvam, V., Lanctot, M.,
  et~al.
\newblock Mastering the game of go with deep neural networks and tree search.
\newblock \emph{nature}, 529\penalty0 (7587):\penalty0 484, 2016.

\bibitem[Sims(2018)]{sims2018efficient}
Sims, C.~R.
\newblock Efficient coding explains the universal law of generalization in
  human perception.
\newblock \emph{Science}, 360\penalty0 (6389):\penalty0 652--656, 2018.

\bibitem[Strens(2000)]{strens2000bayesian}
Strens, M.
\newblock A bayesian framework for reinforcement learning.
\newblock In \emph{ICML}, volume 2000, pp.\  943--950, 2000.

\bibitem[Taylor \& Stone(2009)Taylor and Stone]{taylor2009transfer}
Taylor, M.~E. and Stone, P.
\newblock Transfer learning for reinforcement learning domains: A survey.
\newblock \emph{Journal of Machine Learning Research}, 10\penalty0
  (Jul):\penalty0 1633--1685, 2009.

\bibitem[Tesauro(1995)]{tesauro1995td}
Tesauro, G.
\newblock Temporal {D}ifference {L}earning and {TD-G}ammon.
\newblock \emph{Communications of the ACM}, 38\penalty0 (3):\penalty0 58--68,
  1995.

\bibitem[Tirumala et~al.(2019)Tirumala, Noh, Galashov, Hasenclever, Ahuja,
  Wayne, Pascanu, Teh, and Heess]{tirumala2019exploiting}
Tirumala, D., Noh, H., Galashov, A., Hasenclever, L., Ahuja, A., Wayne, G.,
  Pascanu, R., Teh, Y.~W., and Heess, N.
\newblock Exploiting hierarchy for learning and transfer in kl-regularized rl.
\newblock \emph{arXiv preprint arXiv:1903.07438}, 2019.

\end{thebibliography}
\bibliographystyle{arxiv}

\section{Supplementary Material}
For each supplementary section, refer back to the corresponding section in the full paper for the relevant references.

\subsection{Continuous N-Chain Environment}
\begin{figure}[H]
    \includegraphics[width=1\textwidth]{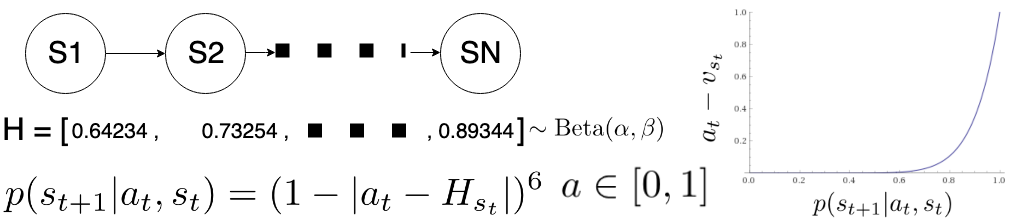}
    \caption{Top: Diagram of the Continuous N-Chain Learning environment. Middle: Example of a set of hidden state values. Right: Graph of the probability of moving to the next state given the absolute distance from the hidden state value and action preformed. Bottom: Function describing this state transition probability.}
\end{figure}

\subsection{Marginal Approximation Method Derivation}
As noted in the main text the basis for the derivation of the continuous action space marginal approximation method is the original MIRL paper.

\textbf{Definition 1 }(stationary distribution over states) The stationary distribution over states (assumed to exist and to be unique) is defined in vector form as $ \mu_{\pi}^T = \lim_{t \rightarrow \infty} v_0^T P_{\pi}^t $ with $v_0$ being an arbitrary vector of probabilities over states at time $t = 0$. The stationary distribution satisfies $\mu_{\pi}(s') = \sum_{s} P_{\pi}(s'|s) \mu_{\pi}(s)$ and therefore is a fixed point under the state transition probabilities $\mu_{\pi}^T = \mu_{\pi}^T P_{\pi}$

\textbf{Definition 2 }(stationary distribution over actions)\textit{ Let $\mu_{\pi}(s)$ be the stationary distribution over states induced by the policy $\pi$. Then the stationary distribution over actions under $\pi$ is defined as $\rho_{\pi}(a) := \sum_{s \in S} \mu_{\pi}(s) \pi(a|s)$}

For the basis $\Pi$ defined by the spherical Gaussian we approximate the marginal action distribution as $\mathcal{N}(\mu_{\rho}, \sigma_{\rho}^2)$ giving the mean and standard deviation of this marginal distribution defined based on the stationary distribution over states and actions with:
\begin{equation*}
\begin{split}
\mu_{\rho} & = \sum_s \mu_{\pi}(s) \mu_s \\
\sigma_{\rho}^2 & = \sum_s p(s) \sigma_s^2 + \sum_s p(s) \big( \mu_s \big)^2 - \bigg( \sum_s p(s) \mu_s \bigg)^2
 \end{split}
\end{equation*}

Because our stationary distribution over states is a fixed point under the state transition probabilities $\mu_{\pi}^T = \mu_{\pi}^T P_{\pi}$, we are able to iteratively update the approximation of our marginal approximation mean ($\hat{\mu}_n$) and standard deviation ($\hat{\sigma}_{\rho}^2$) based on the weighted Gaussian admixture of our previous estimate and the action distribution of the current state:

\textbf{Definition 3 }{Marginal action distribution mean approximation update }{$\hat{\mu}_n = \alpha \mu_n + (1-\alpha) \hat{\mu}_{n-1}$}

\textbf{Definition 4 }{Marginal action distribution standard deviation approximation update }{$\hat{\sigma}_n^2  = \alpha \sigma_n^2 + (1-\alpha) \hat{\sigma}_{n-1}^2 + ( \alpha \mu_n^2 + (1-\alpha) (\hat{\mu}_{n-1})^2) - ( \alpha \mu_n + (1-\alpha) \hat{\mu}_{n-1})^2 $}

We then show that $\lim_{t \rightarrow \infty} \mathcal{N}(\hat{\mu}_n, \hat{\sigma}_{\rho}^2) \rightarrow \mathcal{N}(\mu_{\rho}, \sigma_{\rho}^2)$. This is done by showing that both approximate the mean and variance approach the true mean and variance. This will show that in the limit the approximation of the marginal action distribution approaches the true spherical Gaussian, which itself approximates the true marginal action distribution. This is a fair approximation because the output of the policy network is similarly defined over the spherical Gaussian. 

We can show that the approximate mean/variance approach the true mean and variance because the contribution to our Gaussian admixture which approximate the marginal action distribution eventually approaches the true stationary distribution over actions. This is clear from the definition of the stationary distribution over actions which is induced by the policy $\pi$. Expanding out our approximate mean term to the next step allows us to see where this comes from:

$$\hat{\mu}_n = \alpha \mu_n + (1-\alpha) \hat{\mu}_{n-1}$$
$$\hat{\mu}_{n+1} = \alpha \mu_n + (1-\alpha) (\alpha \mu_n + (1-\alpha) \hat{\mu}_{n-1})$$

From this relation we can see that as $n \rightarrow \infty$ the contribution to the Gaussian admixture from the first state approaches zero, and the overall contribution approaches the average marginal action as defined by the stationary distribution over actions, and the same can be said for the standard deviation as it was constructed with the same method. All that is required is for the $\alpha$ learning rate to approach zero at some point to ensure that the approximation stabilizes, but in practice we do not do this as we do not test agents on an infinite time scale. 

\subsection{KL-RL Derivation}
To show that CLAC can be described as a special case of KL-RL we show the following:
\begin{align}
\mathcal{I}(\pi(a|s)) &= \mathcal{H}(\pi_a(a)) - \mathcal{H}(\pi(a|s)) \\
&= \int_a \pi(a)\log\pi(a) da \\
&- \int_a \pi(a|s)\log\pi(a|s) da\\
&=\int_a\bigg[\int_s p(s|a) ds\bigg]\pi(a)\log\pi(a)da - \\ 
& \int_a\bigg[\int_sp(s)ds\bigg]\pi(a|s)\log\pi(a|s) da\\
&= \int_a\int_s p(s,a)\log\frac{\pi(a)}{\pi(a|s)}dsda\\
&=\mathbb{E}_{s,a}\bigg[\log\frac{\pi(a)}{\pi(a|s)}\bigg]\\
&=\mathbb{E}_{s,a}\bigg[-\log\frac{\pi(a|s)}{\pi(a)}\bigg]
\end{align}
Which is the relation that is used to show that the policy mutual information term can be described as the special case of KL-RL where $\rho(s,a) = \pi_a(a)$
\subsection{Gradients Derivation}
Because of the mathematical connection to the maximum entropy learning objective present in the capacity-limited learning objective, we are able to derive the gradients for training the three DNNs employed by our agent much in the same manner as is used by the original SAC paper. This alteration is particularly straightforward due to our use of the entropy based definition for mutual information. To achieve this derivation, we replace the use of the soft q-function and v-function in SAC with the capacity-limited q-function and v-function, and carry forward this alteration into the remainder of the derivation. 

Taking the capacity-limited value function
\begin{equation*}
\begin{split}
V_{\chi}(s_t) = \EX_{a_t \sim \pi}\big[ Q(s_t,a_t) - \beta(\log\pi_{\mu}(a_t) - \log\pi(a_t|s_t)) \big]
\end{split}
\end{equation*}
With state independent marginal action function $\pi_{\mu}(a_t)$. 
We train the agent to minimize the squared residual error between our capacity-limited value function and $Q_{\kappa}(s_t,a_t)$:
\begin{equation*}
\begin{split}
J_V(\chi) = \EX_{s_t \sim D}\bigg[\dfrac{1}{2}\bigg(V_{\chi}(s_t) - \EX_{a_t \sim \pi_{\phi}} \big[Q_{\kappa}(s_t,a_t) + \beta \big(\log\pi_{\mu}(a_t) - \log\pi_{\phi}(a_t|s_t) \big)\big]\bigg)^2 \bigg]
\end{split}
\end{equation*}
which can be optimized via stochastic gradients with:
\begin{equation*}
\begin{split}
\hat{\nabla}_{\chi} J_V(\chi) = \nabla_{\chi} V_{\chi}(s_t)(V_{\chi}(s_t) - Q_{\kappa}(s_t,a_t) - \beta (\log(\pi_{\mu}(a_t)) - \log(\pi_{\phi}(s_t,a_t)))) 
\end{split}
\end{equation*}
Taking the capacity-limited state-value function:
\begin{equation*}
\begin{split}
\hat{Q}(s_t,a_t) = r(s_t,a_t) + \gamma \EX_{s_{t+1} \sim \rho}[V_{\chi}(s_{t+1})]
\end{split}
\end{equation*}
where $V_{\chi}(s_t)$ is the capacity-limited value function, trained to minimize the Bellman residual: 
\begin{equation*}
\begin{split}
J_Q(\kappa) = \EX_{(s_t,a_t) \sim \mathcal{D}} \bigg[ \dfrac{1}{2}\big(Q_{\kappa}(s_t,a_t) - \hat{Q}_{\kappa}(s_t,a_t)\big)^2 \bigg]
\end{split}
\end{equation*}
again optimized via stochastic gradient descent. 
\begin{equation*}
\begin{split}
\hat{\nabla_{\kappa}} J_Q(\kappa) = \nabla_{\kappa} Q_{\kappa}(s_t,a_t)  (Q_{\kappa}(s_t,a_t) - r(s_t,a_t) - \gamma V_{\hat{\chi}}(s_{t+1}))
\end{split}
\end{equation*}

We begin with the same policy updating method as in SAC, and then replace the soft functions with the capacity-limited functions. This results in a guarantee on improved performance in terms of the capacity-limited value objective. We constrain $\pi \in \Pi$ where $\Pi$ represents the spherical Gaussian family. At each step we update the policy according to
\begin{equation*}
\begin{split}
\pi_{\text{new}} = \argmin_{\pi' \in \Pi} D_{KL}  \Big( \pi'(\cdot|s_t) || \dfrac{\exp(Q^{\pi_{\text{old}}}(s_t, \cdot))}{Z^{\pi_{\text{old}}}(s_t)}  \Big)
\end{split}
\end{equation*}
After employing the policy reparameterization trick used in SAC and KL-RL we can rewrite this objective as:
\begin{equation*}
\begin{split}
J_{\pi}(\phi) = \EX_{s_t \sim D} \big[\EX_{a_t \sim \pi_{\phi}}[\beta \big(log(\pi_{\mu}(a_t)) \\ - log(\pi_{\phi}(a_t|s_t)) \big) - Q_{\kappa}(a_t,s_t)]\big]
\end{split}
\end{equation*}
which gives the gradient as:
\begin{equation*}
\begin{split}
\hat{\nabla_{\phi}} J_{\pi}(\phi) = \nabla_{\phi} \log \pi_{\phi}(a_t|s_t) + \big( \nabla_{a_t} \beta (\log \pi_{\mu}(a_t) \\ - \log \pi_{\phi}(a_t|s_t)) - \nabla_{a_t} Q(a_t,s_t) \big) \nabla_{\phi}f_{\phi}(\eta_t;s_t)
\end{split}
\end{equation*}

\subsection{Automatic Coefficient Update Derivation}
To derive this approach, the SAC method defines the MERL constrained optimization problem as:
\begin{equation}
\begin{split}
\text{max}_{\pi_{0:T}} \EX_{\rho_{\pi}}\bigg[ \sum_{t=0}^T t(s_t,a_t) \bigg] \\ \text{ s.t } \EX_{(s_t,a_t) \sim \rho_{\pi}}[-\log(\pi(a,s))) \leq \mathcal{H}] \text{  } \forall t
\end{split}
\end{equation}
Where $\mathcal{H}$ is the desired minimum target entropy. Using a dynamic programming approach and the recursive definition the MERL soft-Q function, this constrained optimization problem produces the optimal dual variable $\alpha_t^*$ 
\begin{equation}
\alpha_t^* = \argmin_{\alpha_t} \EX_{\alpha_t}[-\alpha_t \log(\pi_t^*(a_t|s_t;\alpha_t)) - \alpha_t \bar{\mathcal{H}}]
\end{equation}
which is calculated in the practical application through gradient descent according to the objective:
\begin{equation}
J(\alpha) = \EX_{a \sim \pi_t}[-\alpha \log \pi_t(a_t|s_t) - \alpha \bar{\mathcal{H}}]
\end{equation}
We begin the derivation of the similar automatic adjustment for the mutual information coefficient by using the entropy definition of the mutual information $\mathcal{I}(\pi(a|s)) = \mathcal{H}(\pi(a)) - \mathcal{H}(\pi(s|a))$ applied to the optimization constraint as follows:
\begin{equation}
\begin{split}
\text{max}_{\pi_{0:T}} \EX_{\rho_{\pi}}\bigg[ \sum_{t=0}^T t(s_t,a_t) \bigg] \\  \text{ s.t }  \EX_{(s_t,a_t) \sim \rho_{\pi}}[-(log(\pi(a) - log(\pi(a,s))) \leq C] \text{  } \forall t
\end{split}
\end{equation}
Because our capacity-limited value function is recursively defined in the same manner as in SAC, and we are using the entropy based definition of the policy mutual information, we can define the optimal dual variable and use a gradient descent method to update this coefficient during learning, giving:
\begin{equation}
J(\beta) = \EX_{a_t \sim \pi_t}[\beta (\log \pi_t(a_t|s_t) - \log \pi_t(a_t|s_t)) + \beta \bar{\mathcal{C}}]
\end{equation}
where $\mathcal{C}$ is the desired maximum policy channel capacity. Thus we can define the automatic mutual information coefficient adjustment version of Algorithm 1, as including the update:
\begin{equation}
\beta \leftarrow \beta - \gamma \hat{\nabla}_{\beta}J(\beta)
\end{equation}
at each gradient step. 

\section{Model Parameters}

The model parameters are kept the same across SAC and CLAC and correspond to those used in the original SAC paper as cited by the stable-baselines repository as the cited source for the code. 

\begin{center}
\begin{tabular}{||c c||} 
\hline
Parameter & Value\\ [0.5ex] 
\hline\hline
optimizer & Adam \\ 
\hline
learning rate & $3 * 10^4$ \\ 
\hline
discount ($\gamma$) & 0.99 \\ 
\hline
replay buffer size & $10^6$ \\ 
\hline
number of hidden layers (all networks) & 2 \\ 
\hline
number of hidden units per layer & 256 \\ 
\hline
number of samples per minibatch  & 256  \\ 
\hline
nonlinearity & ReLU \\ 
\hline
target smoothing coefficient ($\tau$) &  0.005 \\
\hline
target update interval & 1 \\
\hline
gradient steps & 1 \\
\hline
\hline
\end{tabular}
\end{center}

\section{Model Coefficients}
For testing training performance first the range [0-1] is tested in 0.1 spaced increments, then a smaller window around the optimal coefficient is tested repeating this process until there is little performance difference between different coefficients. For all coefficient comparisons 16 training agents are used over the same time scale as reported in the final results. For generalization, the range tested is increased to 0-5 because the models can potentially benefit from larger mutual information or entropy coefficients. In the generalization case the same window based coefficient sampling is used. Other than the mutual information (CLAC) and entropy (SAC) coefficients, all other parameters such as network learning rates and approximation learning rates are kept standard. 

\begin{center}
 \begin{tabular}{||c c c c||} 
 \hline
 Environment & Range Tested & CLAC ($\beta$) & SAC  ($\eta$)\\ [0.5ex] 
 \hline
 \hline
 Continuous N-Chain & 0-5 & 0.5 & 0.5 \\ 
 \hline
 Ant Walker & 0-5 & 0.2 & 0.2 \\
 \hline
 Double Pendulum & 0-5 & 2.0 & 2.0 \\
 \hline
 \hline
\end{tabular}
\end{center}

For all models, coefficients are sampled from 0-5. Firstly, hyper-parameters are sampled in increments of 1 before decreasing the window size until there is no significant difference between different hyper-parameter performance. All hyper-parameters are selecting from the best performing in the environment setting without environment parameter re-sampling. The same hyper-parameters are then used in the setting with environment parameter re-sampling. 

\section{Parameter Sampling}

The following tables describe the environment parameter sampling that is done in the generalization conditions for the continuous n-chain and continuous control environments. The Uni refers to a random uniform re-sampling across the following range. The Beta refers to a beta distributed sampling with (alpha,beta) shape parameters. All pendulum environments randomize the movement force that pushes the balance controlling the arm. In the inverted pendulum and swing-up environments the baseline force is 100 but in the double pendulum it is 200, so in this environment the force is sampled from 180-220. The arm mass and gravity is also sampled from 90-110\% of the original, with the only difference being that in the double pendulum environment the second arm is also re-sampled, independently of the first arm mass. 
\begin{center}
 \begin{tabular}{||c c c||} 
 \hline
 Environment & Parameter & Range\\ 
 \hline \hline
 N-Chain & Hidden Value  & Beta(10,25) \\ 
 \hline
 Double Pendulum  & Arm Mass & Uni [4.5-5.5] \\
 \hline
 Double Pendulum  & 2nd Arm Mass & Uni [4.5-5.5] \\
 \hline
 Double Pendulum & Gravity & Uni [8.82-10.78] \\
 \hline
 Double Pendulum  & Applied Force & Uni [180-220] \\
 \hline
 \hline
\end{tabular}
\end{center}
\begin{center}
 \begin{tabular}{||c c c||} 
 \hline
 Robot Parameter & Random & Extreme \\ [0.8ex] 
 \hline\hline
 Joint Friction & [.475,.525] & [.45,.475] $\cup$  [.525,.55] \\
 \hline 
 Torso Mass & [0.95,1.05] & [.9,.95] $\cup$  [1.05,1.1] \\
 \hline
 Force Power & [.42,.58] & [.38,.42] $\cup$   [.58,.62] \\
 \hline
 Simulated Gravity & [9.31,10.29] & [8.82,9.31] $\cup$  [10.29,10.76] \\
 \hline
\end{tabular}
\end{center}
\end{document}